\pdfoutput=1
\documentclass[11pt]{article}

\usepackage{array}
\usepackage[table]{xcolor}
\usepackage{fix-cm}
\usepackage{auxhook}

\usepackage{times}
\usepackage{latexsym}
\usepackage{float}
\usepackage[T1]{fontenc}
\usepackage{dsfont}

\usepackage[utf8]{inputenc}

\usepackage{microtype}

\usepackage{inconsolata}

\usepackage{graphicx}

\usepackage{arabtex}
\usepackage{utf8}
\usepackage{longtable}
\usepackage{enumitem}
\usepackage{amsmath}
\usepackage{multirow}
\usepackage{booktabs}
\usepackage{listings}
\usepackage{xcolor}
\colorlet{punct}{red!60!black}
\definecolor{background}{HTML}{EEEEEE}
\definecolor{delim}{RGB}{20,105,176}
\colorlet{numb}{magenta!60!black}
\lstdefinelanguage{json}{
    basicstyle=\ttfamily\fontfamily{fvm}\selectfont,
    showstringspaces=false,
    breaklines=true,
    frame=lines,
    backgroundcolor=\color{background},
    literate=
     *{0}{{{\color{numb}0}}}{1}
      {1}{{{\color{numb}1}}}{1}
      {2}{{{\color{numb}2}}}{1}
      {3}{{{\color{numb}3}}}{1}
      {4}{{{\color{numb}4}}}{1}
      {5}{{{\color{numb}5}}}{1}
      {6}{{{\color{numb}6}}}{1}
      {7}{{{\color{numb}7}}}{1}
      {8}{{{\color{numb}8}}}{1}
      {9}{{{\color{numb}9}}}{1}
      {:}{{{\color{punct}{:}}}}{1}
      {,}{{{\color{punct}{,}}}}{1}
      {\{}{{{\color{delim}{\{}}}}{1}
      {\}}{{{\color{delim}{\}}}}}{1}
      {[}{{{\color{delim}{[}}}}{1}
      {]}{{{\color{delim}{]}}}}{1},
}
\usepackage[table]{xcolor}
\usepackage{array}
\usepackage{tabularx} % Required for tabularx environment and X column type
\usepackage{float}

\newcolumntype{L}{>{\raggedright\arraybackslash}X}
\newcolumntype{C}{>{\centering\arraybackslash}X}
\newcolumntype{N}{>{\centering\arraybackslash}p{0.8cm}}
\usepackage[preprint]{acl}
\usepackage{graphicx}
\usepackage{times}
\usepackage{latexsym}
\usepackage{multicol, blindtext}
\usepackage{amsmath}
\usepackage{booktabs}
\usepackage{multirow}
\usepackage[T1]{fontenc}
\usepackage{longtable}
\usepackage{todonotes}
\usepackage[utf8]{inputenc}
\usepackage{microtype}
\usepackage{inconsolata}
\usepackage{tikz}
\usetikzlibrary{arrows,calc,positioning}
\usepackage{graphicx}
\usepackage{float}
\usepackage{amssymb}
\usepackage{stfloats}
\usepackage{enumitem}
\usepackage{tabularx}
\usepackage[table]{xcolor}
\usepackage{float}
\usepackage{array}
\usepackage{xcolor}
\usepackage{colortbl}
\definecolor{mycalmred}{RGB}{240, 160, 160}
\definecolor{mycalmyellow}{RGB}{255, 230, 150}
\definecolor{mycalmgreen}{RGB}{170, 210, 150}
\definecolor{mygreen}{RGB}{198,239,206}
\definecolor{myred}{RGB}{255,199,206}
\definecolor{myredtext}{RGB}{222, 102, 106}
\definecolor{mygreentext}{RGB}{58, 207, 105}
\newtheorem{hyp}{Hypothesis}
\title{Lost in the Mix: Evaluating LLM Understanding of Code-Switched Text}

\author{
 \textbf{Amr Mohamed\textsuperscript{1}$^\dagger$},
 \textbf{Yang Zhang\textsuperscript{2}},
 \textbf{Michalis Vazirgiannis\textsuperscript{1,2}},
 \textbf{Guokan Shang\textsuperscript{1}$^\dagger$}
\\
\\
 \textsuperscript{1}MBZUAI,
 \textsuperscript{2}Ecole Polytechnique
\\
 \small{
$^\dagger$Correspondence: \texttt{\{amr.mohamed, guokan.shang\}@mbzuai.ac.ae}}}
\begin{document}
\maketitle
\begin{abstract}
Code-switching (CSW) is the act of alternating between two or more languages within a single discourse. This phenomenon is widespread in multilingual communities, and increasingly prevalent in online content, where users naturally mix languages in everyday communication. As a result, Large Language Models (LLMs), now central to content processing and generation, are frequently exposed to code-switched inputs. Given their widespread use, it is crucial to understand how LLMs process and reason about such mixed-language text. This paper presents a systematic evaluation of LLM comprehension under code-switching by generating CSW variants of established reasoning and comprehension benchmarks. While degradation is evident when foreign tokens disrupt English text—even under linguistic constraints—embedding English into other languages often improves comprehension. Though prompting yields mixed results, fine-tuning offers a more stable path to degradation mitigation.
\end{abstract}

\section{Introduction}
Code-switching (CSW)—the act of alternating between two or more languages within a single discourse \citep{das-etal-2023-improving,zhang-etal-2023-multilingual,ochieng2024beyond}—is a common phenomenon in multilingual communities \citep{Bullock_Toribio_2009,parekh-etal-2020-understanding,dogruoz-etal-2021-survey}, and increasingly prevalent in online content \citep{kodali2024human}, where users naturally mix languages in everyday informal communications.

Large Language Models (LLMs) have demonstrated remarkable capabilities across a wide range of natural language processing tasks \citep{zhao2023survey}. As they are increasingly used to process and generate content, the widespread availability of code-switched inputs makes it crucial to understand how LLMs reason about such mixed-language data, and whether their multilingual fluency reflects genuine understanding or superficial pattern matching \citep{zhang-etal-2023-multilingual}.
To systematically assess LLMs' handling of such data, we turn to insights from linguistic theories that define the structural constraints governing natural CSW.

Linguistic theories have long studied the structure of CSW text, proposing formal constraints on permissible switch points, such as the Equivalence Constraint Theory (ECT), which posits that switches occur at positions where the surface structures of both languages are grammatically compatible \citep{poplack1978}, and the Matrix Language Frame model (MLF), which distinguishes between a Matrix Language (ML) that provides the grammatical frame of the clause and an Embedded Language (EL) that contributes inserted content without disrupting this structure \citep{myersscotton1993}. These frameworks aim to identify the grammatical boundaries and syntactic compatibility that make CSW possible and natural. While such theories offer testable hypotheses for analyzing CSW, current efforts in synthetic CSW generation often prioritize producing fluent mixed-language text over probing whether LLMs genuinely internalize and apply these structural constraints in their reasoning \citep{pratapa-etal-2018-language,potter-yuan-2024-llm,kuwanto2024linguistics,heredia2025conditioning}.

Despite the availability of well-established linguistic theories, existing evaluation benchmarks fall short of leveraging these insights to assess deeper comprehension in code-switched contexts. Current benchmarks for evaluating the CSW capabilities of language models primarily focus on surface-level tasks 
\citep{khanuja2020gluecos, aguilar-etal-2020-lince, patwa-etal-2020-semeval}. However, they largely overlook the challenge of evaluating deeper reasoning and semantic understanding in mixed-language settings \citep{yadav2024code,gupta2024code,ng2024talking}, leaving a critical gap in assessing the true extent of LLMs’ code-switched comprehension abilities.

\begin{figure*}[t]
    \centering
    \includegraphics[width=\textwidth]{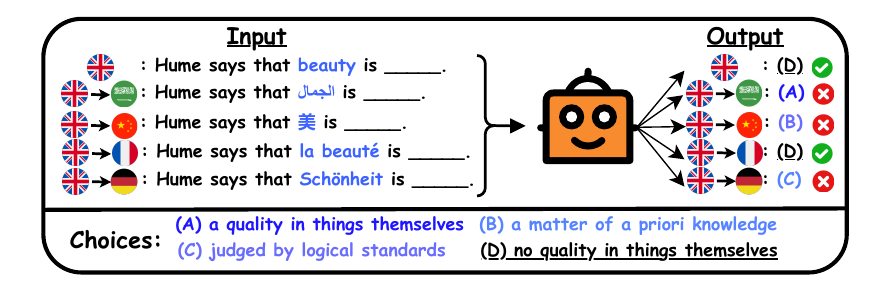}
    \caption{An example illustrating the noun-token CSW methodology from Experiment 1. The figure demonstrates how different embedded languages (Arabic, French, German, Chinese) for the noun ``beauty'' in an English matrix sentence can lead to varied model outputs.}
    \label{fig:illustration} % Or a more specific label if needed
    \label{fig:experiments_illustration}
\end{figure*}

To address these gaps, we introduce a systematic evaluation framework that leverages a constrained, multi‐step LLM pipeline to generate linguistically grounded code‐switched variants of established benchmarks in reading comprehension, multi‐domain knowledge, and natural language inference. Code and data are publicly available\footnote{https://github.com/amr-mohamedd/Lost-in-the-Mix.git}. Our experiments reveal that code‐switching has a nuanced impact on LLM comprehension, influenced by the languages involved and the switching style, as illustrated by the example in Figure \ref{fig:experiments_illustration}. In particular:\\
\noindent $\bullet$ Embedding non‐English tokens into an English matrix language consistently degrades performance, even when the switches follow linguistic constraints, suggesting a structural vulnerability that cannot be explained solely by token-level unfamiliarity.\\
\noindent $\bullet$ Embedding English tokens into non‐English matrix languages often improves comprehension, especially for models with limited proficiency in the matrix language, indicating a facilitative role for English in such contexts.\\
\noindent $\bullet$ While strategic prompting can help some models, it negatively affects others, highlighting inconsistency in controllability; by contrast, fine‐tuning on code‐switched data leads to more stable, albeit partial, performance recovery.

Our work advances the ongoing debate over how LLMs process the \emph{mixed-language content} that now permeates social media, messaging apps, and other corners of the web.  We show that models falter when non-English tokens disrupt an English sentence, yet paradoxically grow more confident when English words are embedded in other languages. This asymmetric behavior reveals a structural imbalance and raises broader concerns about linguistic equity as LLM-generated text is recycled, re-posted, and ultimately re-learned by future models.

\section{Related Work}
\paragraph{Code‐Switching in Language Models.}
Early multilingual encoder-based models (e.g., mBERT \citep{devlin-etal-2019-bert}, XLM-R \citep{conneau-etal-2020-unsupervised}), while effective on monolingual tasks, consistently faltered on code-switched inputs \citep{winata2021multilingual}. This gap spurred specialized methods for mixed-language text, including new architectures and training regimes \citep{winata-etal-2019-code,Liu_Winata_Lin_Xu_Fung_2020,winata-etal-2021-language}. Although existing benchmarks \citep{khanuja2020gluecos} supported these efforts, research predominantly focused on encoder-centric models \citep{winata-etal-2019-code,tan-joty-2021-code,zhu-etal-2023-enhancing}. Consequently, decoder-only architectures, now central to state-of-the-art NLP, have received markedly less scrutiny regarding CSW. While some studies probed adversarial code-mixing in autoregressive models \citep{das2022advcodemix}, meaningful evaluation of such models requires access to high-quality, linguistically coherent code-switched text. This has motivated growing interest in controlled CSW text generation.

\paragraph{Code‐Switched Text Generation.}
Synthetic code-switched text generation plays a critical role in data augmentation and diversification for multilingual language models \citep{pratapa-etal-2018-language,zhang-etal-2023-multilingual}. Methods range from linguistically motivated approaches—such as the Equivalence Constraint Theory (ECT) \citep{poplack1978} and Matrix Language Frame (MLF) model \citep{myersscotton1993}—to heuristic token-level substitutions \citep{Mysln2014CodeswitchingAP,Nguyen02102018,chan-etal-2024-grammatical}. Recent work often relies on word-level aligners to guide borrowing from embedded-language texts while preserving grammatical structure \citep{kuwanto2024linguistics}. Although these techniques aim for token-level accuracy, they overlook the growing capacity of LLMs to perform context-aware, linguistically grounded substitutions. Leveraging this potential, recent studies have explored LLM-based generation using linguistic constraints \citep{kuwanto2024linguistics}, fine-tuning on CSW data \citep{heredia2025conditioning}, or zero-shot prompting \citep{potter-yuan-2024-llm}. Still, challenges remain in controlling switch placement, scaling across language pairs, and conducting robust evaluation. Our work addresses these challenges by leveraging modern LLMs to generate linguistically grounded code-switched text, grounded in established theoretical constraints, to support more rigorous evaluation of model comprehension in mixed-language contexts.

\paragraph{Evaluation of LLM CSW Capabilities.}
LLM CSW evaluation has largely focused on surface-level tasks through benchmarks like GLUECoS \citep{khanuja2020gluecos}, LINCE \citep{aguilar-etal-2020-lince}, and SemEval \citep{patwa-etal-2020-semeval} (e.g., language ID, sentiment, PoS tagging), thus neglecting deeper semantic or reasoning capabilities. Although more recent studies assess CSW sentiment classification \citep{winata2021multilingual}, and question answering \citep{huzaifah-etal-2024-evaluating}, they are limited in scope, emphasizing task-specific metrics over broader comprehension. In contrast, our approach introduces linguistically grounded CSW variants of established comprehension and reasoning tasks, enabling a more rigorous assessment of LLMs' capacity to reason over mixed-language input beyond surface-level performance.

\section{Methodology}
\label{sec:methods}
\subsection{Notations}
\[
\mathcal{B} = \{B_p\}_{p=1}^{P}
\]
be a set of $P$ standard benchmarks. Let
\[
\mathcal{L} = \{l_j\}_{j=1}^{L}
\]
be a set of $L$ languages from which the matrix and embedded languages are selected for code-switched benchmarks generation. Let
\[
\mathcal{M} = \{m_k\}_{k=1}^{K}
\]
be a set of $K$ LLMs.

To evaluate the performance of an LLM $m_k \in \mathcal{M}$ on code-switched text comprehension, we generate a code-switched version of benchmark $B_p \in \mathcal{B}$ using a single matrix language $l_{\text{matrix}} \in \mathcal{L}$ and a set of embedded languages $\mathcal{L}_{\text{embedded}}$, where $\mathcal{L}_{\text{embedded}} \subseteq \mathcal{L}\setminus l_\text{matrix}$  and $|\mathcal{L}_{\text{embedded}}| \ge 1$, which we denote by $B_p^{l_{\text{matrix}}\rightarrow\mathcal{L}_{\text{embedded}}}$.

\subsection{CSW Methods}

To investigate how different CSW strategies affect LLM comprehension, we generate inputs using two distinct approaches: a linguistically grounded \emph{noun-token} method \citep{Poplack1988,muysken2000bilingual,Moyer_2002,chan-etal-2024-grammatical} and a heuristic \emph{ratio-token} method \citep{chan-etal-2024-grammatical}. 

In the noun-token method, we replace nouns in the matrix language text with their aligned counterparts from a parallel sentence in the embedded language. Substitutions are only applied when they preserve grammatical well-formedness according to the Equivalence Constraint Theory and the Matrix Language Frame model, which mandates that the matrix language maintains control over the clause’s morpho-syntactic structure. In contrast, the ratio-token method replaces a ratio of tokens at random, regardless of linguistic structure. This comparison allows us to isolate the role of syntactic and grammatical constraints in LLM comprehension of code-switched text.

\subsection{Code‐Switched Text Generation Approaches}\label{sec:csw_generation_approaches}

Given a parallel corpus, we create code-switched sentences by swapping embedded–language words into matrix–language sentences. To this end, we evaluated two distinct methods for code-switched text generation: an alignment-based method and an LLM-centric method.

\paragraph{Alignment-based method.}
We first align the matrix- and embedded-language sentences with the AWESOME aligner \citep{dou-neubig-2021-word} enhanced by LaBSE embeddings \citep{feng-etal-2022-language}.  
Two variants guide how words are substituted.  
In the \emph{noun-token} variant, we use Stanza POS tagger \citep{qi-etal-2020-stanza} to locate matrix-language nouns and replace each with its aligned counterpart from the embedded-language sentence, prompting Claude 3.5 Sonnet (hereafter \emph{Claude}) to perform the replacements, ensuring that the switch respects the Equivalence Constraint Theory and the Matrix Language Frame model.  
In the \emph{ratio-token} variant, \(\approx20\%\) of aligned tokens are chosen at random and replaced, intentionally relaxing all linguistic constraints to match the setup of \citet{chan-etal-2024-grammatical}.

\paragraph{LLM-centric method.}
Inspired by recent work showing that large language models can fluidly generate code-switched text \citep{potter-yuan-2024-llm}, we let \textit{Claude} perform a two-step procedure.  
First, \textit{Claude} rewrites the matrix-language sentence while inserting masked placeholders at candidate switch points—nouns for the noun-token variant and randomly selected tokens for the ratio-token variant.  
Second, in a subsequent and independent step, \textit{Claude} fills each placeholder with a context-appropriate word taken from the embedded-language sentence, yielding the final code-switched output.

\subsection{Code‐Switching Approach Evaluation}
\label{sec:csw_generation_approaches_eval}

For each embedded language, we assembled a 300-sample test-set, and generated code-switched variants using both approaches from Section \ref{sec:csw_generation_approaches}. GPT-4o then conducted blind, pairwise comparisons under the LLM-as-a-Judge framework \citep{zheng2023judging}, evaluating fluency, depth of mixing, grammatical validity at switch points, and overall coherence. In every case, GPT-4o preferred the two-step LLM-Centric approach, demonstrating its superior capacity to produce high-quality, linguistically coherent code-switched text (See Appendix \ref{app:generation_details} for details on the embedding model, LLM setup, and CSW approach selection and evaluation).

\subsection{Evaluation Metrics}
We evaluate models using three key metrics to capture baseline performance and the effects of code‐switching: accuracy, weighted average accuracy, and accuracy delta.
\paragraph{Accuracy.}
For a model $m_k \in \mathcal{M}$ and benchmark $B'$, whether a monolingual test $B_p \in \mathcal{B}$ or its code-switched variant $B_p^{l_{\text{matrix}}\rightarrow\mathcal{L}_{\text{embedded}}}$, we define accuracy as:
\begin{multline}\label{eq:accuracy} % Added label
  \mathrm{Acc}(m_k, B') = \\
  \frac{1}{|B'|} \sum_{i=1}^{|B'|} \mathds{1}(\mathrm{Correct}(m_k, \text{instance}_i)),
\end{multline}
where $|B'|$ denotes the number of samples in benchmark $B'$, $\text{instance}_i$ is its $i$-th example, and $\mathds{1}(\cdot)$ is the indicator function.

\paragraph{Weighted Average Accuracy.}
To report an aggregate performance measure for a model $m_k$ across multiple benchmarks $\mathcal{B}$, we compute the weighted average accuracy as:
\begin{multline}\label{eq:weighted_accuracy}
  \mathrm{Acc}_{\text{weighted}}(m_k, l_{\text{matrix}}, \mathcal{L}_{\text{embedded}}) =\\
  \frac{\sum_{B_p \in \mathcal{B}} |B_p| \cdot \mathrm{Acc}(m_k, B_p^{l_{\text{matrix}}\rightarrow\mathcal{L}_{\text{embedded}}})}
       {\sum_{B_p \in \mathcal{B}} |B_p|},
\end{multline}

\paragraph{Accuracy Delta ($\Delta\mathrm{Acc}$).}
We quantify the code‐switching impact  by computing the accuracy delta, i.e., the difference between a model’s score on the code‐switched benchmark and its score on the original monolingual benchmark, as:
\begin{multline}\label{eq:degradation} % Added label
  \Delta\mathrm{Acc}(m_k, B_p^{l_{\text{matrix}}\rightarrow\mathcal{L}_{\text{embedded}}}) = \\
  \mathrm{Acc}(m_k, B_p^{l_{\text{matrix}}\rightarrow\mathcal{L}_{\text{embedded}}}) - 
  \mathrm{Acc}(m_k, B_p).
\end{multline}
Positive $\Delta\mathrm{Acc}$ indicates an improvement under code‐switching, negative values a drop.
\section{Experimental Setting}
\paragraph{Languages selection}
We consider a set of languages
\[\mathcal{L} = \{\text{English}, \text{Arabic}, \text{German}, \text{French}, \text{Chinese}\}\]
We hypothesize that this set creates varying degrees of semantic, lexical, and syntactic similarities between the matrix language and the embedded languages set, which may differentially affect the degradation caused by CSW, akin to effects observed in machine translation \citep{guerin2024impact,mohamed2025llm}.

\paragraph{Models selection}
We evaluated LLaMA 3.2 Instruct (3B) and LLaMA 3.1 Instruct (8B, 70B) \citep{grattafiori2024llama}, Qwen 2.5 Instruct (3B, 7B, 72B) \citep{yang2025qwen3}, Mistral 7B Instruct (v0.3) \citep{albert2023mistral}, and ALLaM 7B \citep{bari2024allam}, encompassing a wide range of scales and pretraining curricula. \textit{Allam} currently represents the state-of-the-art in Arabic LLMs, while \textit{Qwen} and \textit{Mistral} excel in Chinese and French, respectively, even as they maintain strong multilingual capabilities. The \textit{Llama} family delivers consistently robust multilingual performance, enabling us to isolate the effects of architecture and model scale on CSW resilience.

\paragraph{Benchmarks selection}
We assess LLM comprehension on three established tasks: \textit{Belebele} \citep{bandarkar2023belebele} for passage-level reading comprehension (with both passages and questions code‐switched), \textit{MMLU}\footnote{https://huggingface.co/datasets/openai/MMMLU} \citep{hendrycks2020measuring} for broad-domain multiple‐choice reasoning (code‐switching applied to questions), and \textit{XNLI} \citep{conneau-etal-2018-xnli} natural language inference (both premise and hypothesis code‐switched). To ensure consistent, scalable evaluation across models, we used and adapted EleutherAI's Language Model Evaluation Harness \citep{eval-harness} for our code‐switched variants.
\section{Experiments}
\label{sec:experiments}
\subsection{Experiment 1: Linguistically motivated CSW}
\paragraph{Setup} We use English as the matrix language $l_{\text{matrix}}$, and perform CSW on the benchmarks with each language in $\mathcal{L} \setminus l_{\text{matrix}}$ as the embedded language separately, using the noun-token CSW method, and compare the performance of the code-switched benchmarks with the original English benchmarks.
\begin{hyp}[H\ref{hyp:first}]\label{hyp:first} We hypothesize that LLM performance on code-switched benchmarks degrades in proportion to the linguistic distance between the matrix and embedded languages.\end{hyp}

\begin{figure*}[t]
  \center
    \includegraphics[width=\textwidth]{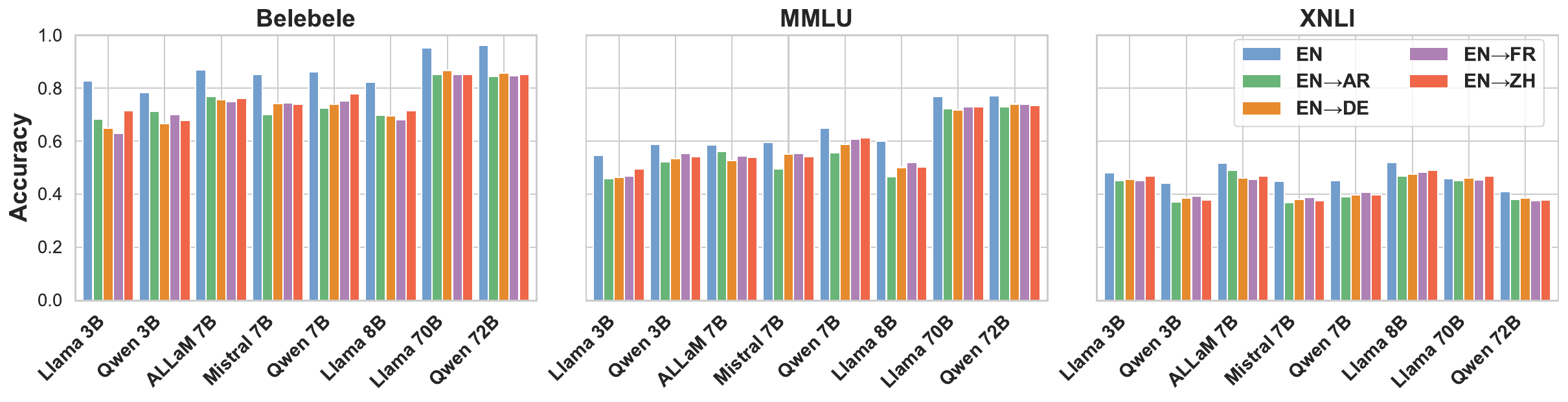}
    \caption{Comparison of LLM accuracy on monolingual English versions of \textit{Belebele}, \textit{MMLU}, and \textit{XNLI} benchmarks (baseline) versus their noun-token code-switched counterparts. English serves as the matrix language, with Arabic (EN$\rightarrow$AR), French (EN$\rightarrow$FR), German (EN$\rightarrow$DE), and Chinese (EN$\rightarrow$ZH) as embedded languages.}
    \label{fig:noun-token}
\end{figure*}

\paragraph{Results} Table \ref{tab:table1-finalglobal} and Figure \ref{fig:noun-token} show consistent drops in LLM performance on noun-token code-switched benchmarks compared to their English versions. The extent of degradation varied by embedded language and model. For example, LLaMA-70B's weighted average accuracy declined from 0.70 (English) to 0.66 on EN$\rightarrow$AR/EN$\rightarrow$DE ($\Delta \approx -0.04$) and 0.67 on EN$\rightarrow$ZH ($\Delta \approx -0.03$).

Mistral-7B showed minimal loss on EN$\rightarrow$FR ($\Delta \approx -0.01$), and ALLaM-7B retained relatively strong performance on EN$\rightarrow$AR ($\Delta \approx -0.06$). Qwen models exhibited consistent degradation across languages (e.g., Qwen-7B: $\Delta \approx -0.03$ to $-0.06$), with larger models achieving better absolute scores but similar relative drops. These trends held across all three tasks, underscoring both the general difficulty of CSW and the role of language-specific model strengths.

\begin{table}[t]
  \small
  \centering
    \begin{tabularx}{\linewidth}{l !{\vrule width 1pt} C !{\vrule width 1pt} C !{\vrule width 1pt} C !{\vrule width 1pt} C !{\vrule width 1pt} N}
  \hline
  \textbf{Model}
    & \textbf{EN$\rightarrow$AR}
    & \textbf{EN$\rightarrow$DE}
    & \textbf{EN$\rightarrow$FR}
    & \textbf{EN$\rightarrow$ZH}
    & \textbf{EN} \\
  \hline
  \textbf{Llama 3B}     
    & \cellcolor{myred!61!mygreen}0.47 
    & \cellcolor{myred!61!mygreen}0.47 
    & \cellcolor{myred!61!mygreen}0.47 
    & \cellcolor{myred!54!mygreen}0.50 
    & \cellcolor{myred!44!mygreen}\textbf{0.54} \\
  \textbf{Qwen 3B}       
    & \cellcolor{myred!56!mygreen}0.49 
    & \cellcolor{myred!54!mygreen}0.50 
    & \cellcolor{myred!46!mygreen}0.52 
    & \cellcolor{myred!51!mygreen}0.51 
    & \cellcolor{myred!39!mygreen}\textbf{0.56} \\
  \textbf{Allam 7B} 
    & \cellcolor{myred!41!mygreen}0.55 
    & \cellcolor{myred!49!mygreen}0.52 
    & \cellcolor{myred!46!mygreen}0.53 
    & \cellcolor{myred!46!mygreen}0.53 
    & \cellcolor{myred!34!mygreen}\textbf{0.58} \\
  \textbf{Mistral 7B}  
    & \cellcolor{myred!61!mygreen}0.47 
    & \cellcolor{myred!49!mygreen}0.52 
    & \cellcolor{myred!49!mygreen}0.52 
    & \cellcolor{myred!51!mygreen}0.51 
    & \cellcolor{myred!37!mygreen}\textbf{0.57} \\
  \textbf{Qwen 7B}
    & \cellcolor{myred!49!mygreen}0.52 
    & \cellcolor{myred!41!mygreen}0.55 
    & \cellcolor{myred!39!mygreen}0.56 
    & \cellcolor{myred!37!mygreen}0.57 
    & \cellcolor{myred!27!mygreen}\textbf{0.61} \\
  \textbf{Llama 8B}     
    & \cellcolor{myred!59!mygreen}0.48 
    & \cellcolor{myred!51!mygreen}0.51 
    & \cellcolor{myred!49!mygreen}0.52 
    & \cellcolor{myred!51!mygreen}0.51 
    & \cellcolor{myred!32!mygreen}\textbf{0.59} \\
  \textbf{Llama 70B}    
    & \cellcolor{myred!15!mygreen}0.66 
    & \cellcolor{myred!15!mygreen}0.66 
    & \cellcolor{myred!12!mygreen}0.67 
    & \cellcolor{myred!12!mygreen}0.67 
    & \cellcolor{myred!5!mygreen}\textbf{0.70} \\
  \textbf{Qwen 72B}  
    & \cellcolor{myred!17!mygreen}0.65 
    & \cellcolor{myred!15!mygreen}0.66 
    & \cellcolor{myred!17!mygreen}0.65 
    & \cellcolor{myred!17!mygreen}0.65 
    & \cellcolor{myred!7!mygreen}\textbf{0.69} \\
  \hline
  \end{tabularx}
  \caption{Weighted average accuracy of selected LLMs on noun-token code-switched benchmarks (EN$\rightarrow$ AR, EN$\rightarrow$DE, EN$\rightarrow$FR, EN$\rightarrow$ZH) compared to the monolingual English baseline. Cell colors indicate relative performance from highest (green) to lowest (red). The highest scores are indicated in \textbf{bold}.}
  \label{tab:table1-finalglobal}
\end{table}
\subsection{Experiment 2: Non-linguistically motivated CSW}
\paragraph{Setup} In this experiment, we retain the experimental framework of Experiment 1, replacing the linguistically motivated noun‐token CSW method with the ratio‐token method.

\begin{hyp}[H\ref{hyp:second}]\label{hyp:second} We hypothesize that non-linguistically motivated CSW leads to sharper performance degradation in LLMs than that observed on linguistically motivated CSW, as such input is less likely to align with patterns encountered during pre-training.\end{hyp}

\paragraph{Results} 
Results are show in Table~\ref{table:ratio-token}. All models exhibited a decline in weighted average accuracy, consistent with the patterns observed in Experiment~1. The extent of degradation varied with model size and language pairing. Smaller models experienced the most pronounced drops; for example, \textit{Llama 3B} decreased from 0.54 (EN) to 0.43 on EN$\rightarrow$DE ($\Delta = -0.11$) and to 0.47 on EN$\rightarrow$AR ($\Delta = -0.07$). In contrast, \textit{Llama 70B} showed minimal degradation, with weighted average accuracy decreasing from 0.70 to 0.68 across all embedded languages ($\Delta \approx -0.02$). Language-specific resilience was also observed. \textit{Allam 7B} and \textit{Mistral 7B} relatively strong performance on EN$\rightarrow$AR on EN$\rightarrow$FR, respectively. \textit{Qwen 7B} exhibited consistent, moderate degradation, decreasing from 0.61 to a range of 0.53–0.57 depending on the embedded language ($\Delta = -0.08$ to $-0.04$).% Overall, the performance degradation observed under ratio-token code-switching was broadly comparable to that of the noun-token setting, with no systematic advantage for either style across models or language pairs.
\begin{table}[t]
  \small
  \centering
    \begin{tabularx}{\linewidth}{l !{\vrule width 1pt} C !{\vrule width 1pt} C !{\vrule width 1pt} C !{\vrule width 1pt} C !{\vrule width 1pt} N}
  \hline
  \textbf{Model}
    & \textbf{EN$\rightarrow$AR}
    & \textbf{EN$\rightarrow$DE}
    & \textbf{EN$\rightarrow$FR}
    & \textbf{EN$\rightarrow$ZH}
    & \textbf{EN} \\
  \hline
  \textbf{Llama 3B}     
    & \cellcolor{myred!61!mygreen}0.47  
    & \cellcolor{myred!71!mygreen}0.43
    & \cellcolor{myred!63!mygreen}0.46
    & \cellcolor{myred!51!mygreen}0.51
    & \cellcolor{myred!44!mygreen}\textbf{0.54} \\
  \textbf{Qwen 3B}       
    & \cellcolor{myred!55!mygreen}0.50
    & \cellcolor{myred!51!mygreen}0.51
    & \cellcolor{myred!49!mygreen}0.52
    & \cellcolor{myred!51!mygreen}0.51
    & \cellcolor{myred!39!mygreen}\textbf{0.56} \\
  \textbf{Allam 7B} 
    & \cellcolor{myred!36!mygreen}0.56
    & \cellcolor{myred!51!mygreen}0.51
    & \cellcolor{myred!46!mygreen}0.53
    & \cellcolor{myred!44!mygreen}0.54
    & \cellcolor{myred!34!mygreen}\textbf{0.58} \\
  \textbf{Mistral 7B}  
    & \cellcolor{myred!57!mygreen}0.49
    & \cellcolor{myred!49!mygreen}0.52
    & \cellcolor{myred!46!mygreen}0.53
    & \cellcolor{myred!49!mygreen}0.52
    & \cellcolor{myred!37!mygreen}\textbf{0.57} \\
  \textbf{Qwen 7B}
    & \cellcolor{myred!46!mygreen}0.53
    & \cellcolor{myred!41!mygreen}0.55
    & \cellcolor{myred!39!mygreen}0.56
    & \cellcolor{myred!37!mygreen}0.57
    & \cellcolor{myred!27!mygreen}\textbf{0.61} \\
  \textbf{Llama 8B}     
    & \cellcolor{myred!54!mygreen}0.50
    & \cellcolor{myred!49!mygreen}0.52
    & \cellcolor{myred!46!mygreen}0.53
    & \cellcolor{myred!44!mygreen}0.54
    & \cellcolor{myred!32!mygreen}\textbf{0.59} \\
  \textbf{Llama 70B}    
    & \cellcolor{myred!10!mygreen}0.68
    & \cellcolor{myred!12!mygreen}0.67
    & \cellcolor{myred!10!mygreen}0.68
    & \cellcolor{myred!10!mygreen}0.68
    & \cellcolor{myred!5!mygreen}\textbf{0.70} \\
  \textbf{Qwen 72B}  
    & \cellcolor{myred!15!mygreen}0.66
    & \cellcolor{myred!15!mygreen}0.66
    & \cellcolor{myred!15!mygreen}0.66
    & \cellcolor{myred!15!mygreen}0.66
    & \cellcolor{myred!7!mygreen}\textbf{0.69} \\
  \hline
  \end{tabularx}
  \caption{Weighted average accuracy of selected LLMs on ratio-token code-switched benchmarks (EN$\rightarrow$ AR, EN$\rightarrow$DE, EN$\rightarrow$FR, EN$\rightarrow$ZH) compared to the monolingual English baseline. Cell colors indicate relative performance from highest (green) to lowest (red). The highest scores are indicated in \textbf{bold}.}
  \label{table:ratio-token}
\end{table}

\section{Ablations}
Building on Section~\ref{sec:experiments}, which found comparable degradation from noun-token and ratio-token CSW, we proceed with ablation studies using exclusively the noun-token method.
\subsection{English as an embedded language}
To assess whether embedding English improves comprehension in other matrix languages, we reversed the language roles from the main experiments, using each language in $\mathcal{L} \setminus l_{\text{matrix}}$ as the matrix language, and English as the sole embedded language. We generated code-switched versions ($B_p^{l_{\text{matrix}} \rightarrow \{\text{English}\}}$) of the \textit{Belebele}, \textit{MMLU}, and \textit{XNLI} benchmarks. By comparing model performance on these variants against their original monolingual counterparts, we aimed to assess any comprehension enhancement attributable to the embedded English words.

\renewcommand{\arraystretch}{1.1}   % scales row height by 1.1
\begin{table}[t]
  \small
  \centering
  \begin{tabular}{%
    l!{\vrule width .6pt}
    @{\,}rr@{\;}!{\vrule width .5pt}
    @{\,}rr@{\;}!{\vrule width .5pt}
    @{\,}rr@{\;}!{\vrule width .5pt}
    @{\,}rr
  }
  \hline
  \noalign{\vskip 2pt}                   % extra space above the first header
  \multirow{2}{*}{\textbf{Model}}
    & \multicolumn{2}{c}{\textbf{AR$\rightarrow$EN}}
    & \multicolumn{2}{c}{\textbf{DE$\rightarrow$EN}}
    & \multicolumn{2}{c}{\textbf{FR$\rightarrow$EN}}
    & \multicolumn{2}{c}{\textbf{ZH$\rightarrow$EN}} \\
  \noalign{\vskip 2pt}                   % extra space below the first header
  \cline{2-9}
  \noalign{\vskip 1pt}                   % extra space above the sub-column names
    & Orig & CSW
    & Orig & CSW
    & Orig & CSW
    & Orig & CSW \\
  \hline
  \textbf{Llama 3B}
    & 0.37 & \textbf{0.45}
    & 0.35 & \textbf{0.38}
    & 0.43 & \textbf{0.45}
    & 0.42 & \textbf{0.47} \\
  \textbf{Qwen 3B}
    & 0.40 & \textbf{0.48}
    & 0.49 & \textbf{0.52}
    & 0.50 & \textbf{0.53}
    & \textit{0.48} & \textit{0.48} \\
  
  \textbf{Allam 7B}
    & 0.51 & \textbf{0.52}
    & 0.39 & \textbf{0.43}
    & 0.49 & \textbf{0.52}
    & 0.44 & \textbf{0.51} \\
  
  \textbf{Mistral 7B}
    & 0.35 & \textbf{0.48}
    & 0.50 & \textbf{0.54}
    & 0.52 & \textbf{0.55}
    & 0.46 & \textbf{0.53} \\
  
  \textbf{Qwen 7B}
    & 0.47 & \textbf{0.52}
    & 0.51 & \textbf{0.53}
    & 0.56 & \textbf{0.57}
    & \textbf{0.56} & 0.55 \\
  
  \textbf{Llama 8B}
    & 0.38 & \textbf{0.44}
    & \textit{0.50} & \textit{0.50}
    & 0.50 & \textbf{0.52}
    & 0.49 & \textbf{0.53} \\
  
  \textbf{Llama 70B}
    & 0.61 & \textbf{0.66}
    & 0.67 & \textbf{0.67}
    & \textit{0.68} & \textit{0.68}
    & 0.64 & \textbf{0.66} \\
  
  \textbf{Qwen 72B}
    & 0.63 & \textbf{0.66}
    & \textit{0.68} & \textit{0.68}
    & \textit{0.68} & \textit{0.68}
    & \textit{0.66} & \textit{0.66} \\
  \hline
  \end{tabular}
    \caption{Weighted average accuracy of LLMs on monolingual (Orig) versus English-embedded code-switched (CSW) benchmarks across Arabic, German, French, and Chinese, rounded to two decimals. \textbf{Bold} indicates the higher score in each Orig/CSW pair. \textit{Italic} indicates instances where performance did not change between the original and code-switched versions.}
  \label{table:l-en}
  \end{table}
Results are presented in Table~\ref{table:l-en}. Embedding English into lower-resource matrix languages often improved model performance or, at minimum, avoided large degradations. Gains were especially prominent when models lacked proficiency in the matrix language. For instance, \textit{Mistral 7B}'s weighted average accuracy in Arabic rose from 0.35 to 0.48 ($\Delta=+0.13$), while its score in Chinese increased by +0.07 points. In contrast, when models already demonstrated strong matrix language proficiency, improvements were minimal or absent. \textit{Allam 7B} (Arabic) and \textit{Mistral 7B} (French) saw gains of only +0.01 and +0.03, respectively. High-performing models such as \textit{Llama 70B} and \textit{Qwen 72B} showed no change in several settings. Only one case showed a minor drop: \textit{Qwen 7B} on Chinese ($\Delta \approx -0.01$). This suggests that embedded English may introduce interference when matrix language representations are already strong.
\subsection{When Code-Switching Goes Extreme}
To assess performance under more complex multilingual mixing, an "extreme" CSW experiment was conducted on the \textit{MMLU} benchmark. English served as the matrix language, with nouns code-switched using three distinct embedded languages sets: \\\textbf{Setting 1} featured a non-Latin script pair ($\mathcal{L}_{\text{embedded}} = \{\text{Arabic,Chinese}\}$), \\\textbf{Setting 2} used a Latin script pair ($\mathcal{L}_{\text{embedded}} = \{\text{French,German}\}$), and \\\textbf{Setting 3} combined all four languages ($\mathcal{L}_{\text{embedded}} = \{\text{Arabic,Chinese,French,German}\}$).\\For generating the code-switched text across these settings, Claude was, additionally, prompted to borrow words evenly from the specified embedded languages for each instance.
\begin{table}[t]
  \small
  \centering
  \setlength{\tabcolsep}{0pt}
  \begin{tabularx}{\linewidth}{%
    >{\raggedright\arraybackslash}X!{\vrule width .6pt}
    >{\centering\arraybackslash}X!{\vrule width .5pt}
    >{\centering\arraybackslash}X!{\vrule width .5pt}
    >{\centering\arraybackslash}X!{\vrule width .5pt}
    >{\centering\arraybackslash}X
  }
  \hline
  \textbf{Model}
    & \textbf{Setting 1} % val 0.48 -> P=(0.48-0.31)/(0.77-0.31)=0.17/0.46=0.37 -> 63% red
    & \textbf{Setting 2} % val 0.46 -> P=(0.46-0.31)/(0.77-0.31)=0.15/0.46=0.33 -> 67% red
    & \textbf{Setting 3} % val 0.47 -> P=(0.47-0.31)/(0.77-0.31)=0.16/0.46=0.35 -> 65% red
    & \textbf{EN} \\ % val 0.55 -> P=(0.55-0.31)/(0.77-0.31)=0.24/0.46=0.52 -> 48% red
  \hline
  \textbf{Llama 3B}  
    & \cellcolor{myred!63!mygreen}0.48 
    & \cellcolor{myred!67!mygreen}0.46 
    & \cellcolor{myred!65!mygreen}0.47 
    & \cellcolor{myred!48!mygreen}\textbf{0.55} \\
  \textbf{Qwen 3B}   % 0.54 -> P=0.50 -> 50% red; 0.55 -> P=0.52 -> 48% red; 0.53 -> P=0.48 -> 52% red; 0.59 -> P=0.61 -> 39% red
    & \cellcolor{myred!50!mygreen}0.54 
    & \cellcolor{myred!48!mygreen}0.55 
    & \cellcolor{myred!52!mygreen}0.53 
    & \cellcolor{myred!39!mygreen}\textbf{0.59} \\
  \textbf{Allam 7B } % 0.56 -> P=0.54 -> 46% red; 0.54 -> P=0.50 -> 50% red; 0.54 -> P=0.50 -> 50% red; 0.58 -> P=0.59 -> 41% red
    & \cellcolor{myred!46!mygreen}0.56 
    & \cellcolor{myred!50!mygreen}0.54 
    & \cellcolor{myred!50!mygreen}0.54 
    & \cellcolor{myred!41!mygreen}\textbf{0.58} \\
  \textbf{Mistral 7B}% 0.53 -> P=0.48 -> 52% red; 0.56 -> P=0.54 -> 46% red; 0.55 -> P=0.52 -> 48% red; 0.59 -> P=0.61 -> 39% red
    & \cellcolor{myred!52!mygreen}0.53 
    & \cellcolor{myred!46!mygreen}0.56 
    & \cellcolor{myred!48!mygreen}0.55 
    & \cellcolor{myred!39!mygreen}\textbf{0.59} \\
  \textbf{Qwen 7B}   % 0.58 -> P=0.59 -> 41% red; 0.60 -> P=0.63 -> 37% red; 0.59 -> P=0.61 -> 39% red; 0.65 -> P=0.74 -> 26% red
    & \cellcolor{myred!41!mygreen}0.58 
    & \cellcolor{myred!37!mygreen}0.60 
    & \cellcolor{myred!39!mygreen}0.59 
    & \cellcolor{myred!26!mygreen}\textbf{0.65} \\
  \textbf{Llama 8B}  % 0.49 -> P=0.39 -> 61% red; 0.51 -> P=0.43 -> 57% red; 0.49 -> P=0.39 -> 61% red; 0.60 -> P=0.63 -> 37% red
    & \cellcolor{myred!61!mygreen}0.49 
    & \cellcolor{myred!57!mygreen}0.51 
    & \cellcolor{myred!61!mygreen}0.49 
    & \cellcolor{myred!37!mygreen}\textbf{0.60} \\
  \textbf{Llama 70B} % 0.72 -> P=0.89 -> 11% red; 0.70 -> P=0.85 -> 15% red; 0.70 -> P=0.85 -> 15% red; 0.77 -> P=1.00 -> 0% red (all green)
    & \cellcolor{myred!11!mygreen}0.72 
    & \cellcolor{myred!15!mygreen}0.70 
    & \cellcolor{myred!15!mygreen}0.70 
    & \cellcolor{myred!0!mygreen}\textbf{0.77} \\ 
  \textbf{Qwen 72B}  % 0.74 -> P=0.93 -> 7% red; 0.74 -> P=0.93 -> 7% red; 0.73 -> P=0.91 -> 9% red; 0.77 -> P=1.00 -> 0% red (all green)
    & \cellcolor{myred!7!mygreen}0.74 
    & \cellcolor{myred!7!mygreen}0.74 
    & \cellcolor{myred!9!mygreen}0.73 
    & \cellcolor{myred!0!mygreen}\textbf{0.77} \\
  \hline
  \end{tabularx}
  \caption{\textit{MMLU} accuracy for extreme CSW with English as the matrix language and the embedded languages being Arabic and Chinese (Setting 1), French and German (Setting 2), and Arabic, Chinese, French, and German (Setting 3), alongside the monolingual English baseline. The highest scores are indicated in bold.}

  \label{tab:extreme_csw}
\end{table}
Table \ref{tab:extreme_csw} demonstrates that all models experience a decline in \textit{MMLU} accuracy under extreme code‐switching relative to the monolingual English baseline. For example, \textit{Llama 70B}’s score decreases from 0.77 to between 0.70 and 0.72, and \textit{Qwen 72B}’s from 0.77 to 0.73–0.74. Analyzing language-script effects by comparing the non‐Latin mix (Setting 1) against the Latin mix (Setting 2) reveals no uniform penalty for non‐Latin scripts. \textit{Allam 7B} achieves a higher accuracy with the non‐Latin pair (0.56 vs. 0.54), whereas \textit{Mistral 7B} performs better with the Latin pair (0.56 vs. 0.53). Moreover, extending the embedded set to all four languages (Setting 3) does not invariably yield the lowest scores, while \textit{Llama 70B} (0.70) and \textit{Qwen 72B} (0.73) record their minima in Setting 3, other models exhibit accuracies intermediate between those in Settings 1 and 2.
\section{Mitigation strategies}
To mitigate the performance declines induced by CSW, we investigate two strategies: a prompt-based approach, which prepends explicit instructions to code-switched inputs, and a model-based approach, which fine-tunes LLMs on synthetic CSW data. 
\subsection{Prompt-based Mitigation}
Each noun-token code-switched benchmark instance was prepended with an explicit instruction indicating that the input involves English mixed with an embedded language. Further details on the prompts used per benchmark are provided in Appendix \ref{app:prompt_mitigation}.
\begin{table}[t]
  \small
  \centering
    \begin{tabularx}{\linewidth}{l !{\vrule width 1pt} C !{\vrule width 1pt} C !{\vrule width 1pt} C !{\vrule width 1pt} C !{\vrule width 1pt} N}
  \hline
  \textbf{Model}
  & \textbf{EN$\rightarrow$AR}
  & \textbf{EN$\rightarrow$DE}
  & \textbf{EN$\rightarrow$FR}
  & \textbf{EN$\rightarrow$ZH}
  & \textbf{EN} \\
  \hline
  \textbf{Llama 3B}
  & \cellcolor{myred!100!mygreen}0.31
  & \cellcolor{myred!93!mygreen}0.34
  & \cellcolor{myred!98!mygreen}0.32
  & \cellcolor{myred!98!mygreen}0.32
  & \cellcolor{myred!44!mygreen}\textbf{0.54} \\
  \textbf{Qwen 3B}
  & \cellcolor{myred!49!mygreen}0.51
  & \cellcolor{myred!46!mygreen}0.53
  & \cellcolor{myred!44!mygreen}0.54
  & \cellcolor{myred!46!mygreen}0.53
  & \cellcolor{myred!39!mygreen}\textbf{0.56} \\
\textbf{Allam 7B}
  & \cellcolor{myred!39!mygreen}0.56
  & \cellcolor{myred!46!mygreen}0.53
  & \cellcolor{myred!44!mygreen}0.54
  & \cellcolor{myred!46!mygreen}0.53
  & \cellcolor{myred!34!mygreen}\textbf{0.58} \\
  \textbf{Mistral 7B}
  & \cellcolor{myred!63!mygreen}0.46
  & \cellcolor{myred!54!mygreen}0.50
  & \cellcolor{myred!54!mygreen}0.50
  & \cellcolor{myred!54!mygreen}0.50
  & \cellcolor{myred!37!mygreen}\textbf{0.57} \\
  \textbf{Qwen 7B}
  & \cellcolor{myred!44!mygreen}0.54
  & \cellcolor{myred!39!mygreen}0.56
  & \cellcolor{myred!34!mygreen}0.58
  & \cellcolor{myred!32!mygreen}0.59
  & \cellcolor{myred!27!mygreen}\textbf{0.61} \\
  \textbf{Llama 8B}
  & \cellcolor{myred!76!mygreen}0.41
  & \cellcolor{myred!61!mygreen}0.47
  & \cellcolor{myred!59!mygreen}0.48
  & \cellcolor{myred!61!mygreen}0.47
  & \cellcolor{myred!32!mygreen}\textbf{0.59} \\
  \textbf{Llama 70B}
  & \cellcolor{myred!46!mygreen}0.53
  & \cellcolor{myred!46!mygreen}0.53
  & \cellcolor{myred!20!mygreen}0.64
  & \cellcolor{myred!54!mygreen}0.50
  & \cellcolor{myred!5!mygreen}\textbf{0.70} \\
  \textbf{Qwen 72B}
  & \cellcolor{myred!5!mygreen}0.70
  & \cellcolor{myred!2!mygreen}0.71
  & \cellcolor{myred!2!mygreen}0.71
  & \cellcolor{myred!0!mygreen}\textbf{0.72}
  & \cellcolor{myred!7!mygreen}0.69 \\
  \hline
  \end{tabularx}
  \caption{Impact of an instructional prompt on LLM weighted average accuracy for noun-token code-switched benchmarks. English serves as the matrix language, with results shown for various embedded languages. The highest scores are indicated in \textbf{bold}}
  \label{tab:prompting_results_super_super_global}
\end{table}

The results of the prompt-based mitigation approach, presented in Table \ref{tab:prompting_results_super_super_global}, show considerable variation across models when compared to unprompted noun-token CSW (Table \ref{tab:table1-finalglobal}). For some models, most notably the \textit{Qwen} family, the addition of an explicit instruction led to consistent performance gains. \textit{Qwen 72B} improved across all language pairs, most remarkably surpassing its monolingual English weighted average accuracy (EN$\rightarrow$ZH: 0.72 vs. EN: 0.69). Similarly, \textit{Qwen 7B} also benefited, with EN$\rightarrow$ZH improving from 0.57 to 0.59 ($\Delta=+0.02$). \textit{Allam 7B} exhibited minor improvements as well, such as EN$\rightarrow$AR increasing from 0.55 to 0.56 ($\Delta=+0.01$).

Conversely, for other models, particularly the \textit{Llama} family and \textit{Mistral 7B}, the prompt-based strategy was frequently detrimental. \textit{Llama 8B} saw weighted average accuracy declines across all embedded languages (e.g., EN$\rightarrow$FR dropped from 0.52 to 0.48, $\Delta=-0.04$). More substantial drops were observed for \textit{Llama 70B}, especially on EN$\rightarrow$AR and EN$\rightarrow$ZH, where performance fell by 13 and 17 points respectively. \textit{Llama 3B} and \textit{Mistral 7B} similarly exhibited declines (e.g., \textit{Llama 3B} EN$\rightarrow$AR:a $\Delta=-0.16$).

\subsection{Model-based Mitigation}
\label{sec:model_mitigation}
Directly fine-tuning LLMs on code-switched text presents another avenue for mitigation. For this, \textit{Llama 8B} was selected, primarily due to its limited responsiveness to prompting within its size category. A parallel corpus of TED Talk transcripts \citep{qi-etal-2018-pre} spanning English, Arabic, Chinese, French, and German was utilized. The instruction-tuning dataset was constructed by first selecting samples from the parallel corpus where the English sentence length was greater than 70 words. This filtering yielded approximately 3,650 pairs per language combination. Noun-token CSW, with English as a matrix language, was then applied to these, resulting in an instruction-tuning dataset of approximately 14,600 training samples. The instruction required the model to generate the code-switched text from the original English and embedded-language sentences, using five distinct prompt templates to ensure instructions diversity (further details in Appendix \ref{app:ift_details_full}).
\begin{figure}[t]
  \centering
  \includegraphics[width=\columnwidth]{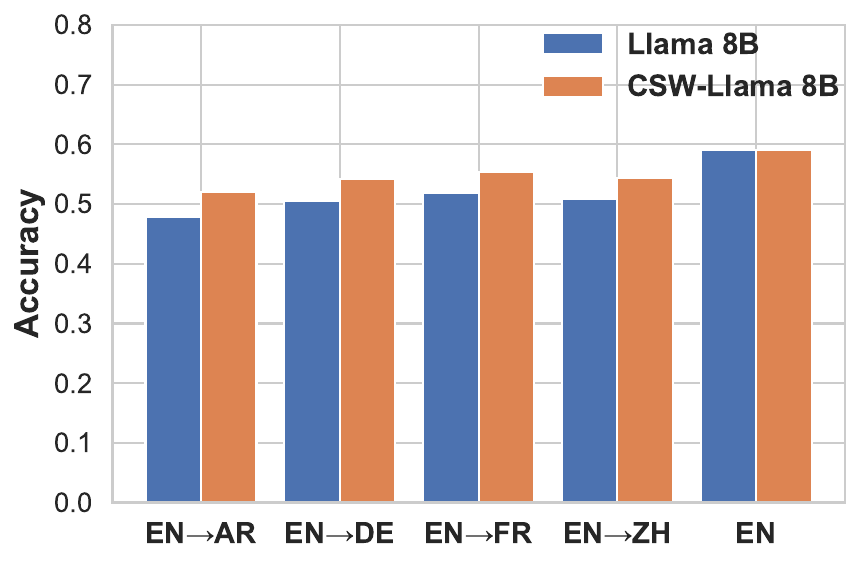}
    \caption{Comparison of \textit{Llama 8B} and its instruction-tuned variant (\textit{CSW-Llama 8B}) on monolingual English benchmarks (\textit{Belebele}, \textit{MMLU}, and \textit{XNLI}) versus their noun-token code-switched counterparts. English serves as the matrix language, with Arabic, French, German, and Chinese, as embedded languages.}
  \label{fig:mitigation-IFT}
\end{figure}

The impact of this instruction fine-tuning is illustrated in Figure~\ref{fig:mitigation-IFT}. The baseline \textit{Llama 8B} model achieved an English-only weighted average accuracy of 0.59 on the combined benchmarks. Introducing noun-token CSW without fine-tuning resulted in a weighted average accuracy reduction of up to 0.11 points, depending on the embedded language. After fine-tuning on the code-switched corpus (yielding \textit{CSW-Llama 8B}), a partial recovery of performance was observed. The most significant improvement was for the EN$\rightarrow$AR setting, where the weighted average accuracy increased by +0.04 points over the baseline. The smallest gain was for EN$\rightarrow$FR, with an increase of +0.03 points.
\section{Discussion and Conclusion} % Or just \section{Discussion} if you have a separate Conclusion
\label{sec:discussion}

As LLMs increasingly process multilingual and mixed-language inputs, understanding their comprehension limits is paramount. This study systematically evaluated LLM performance on code-switched text, yielding multifaceted insights into information processing under these conditions. Our findings reveal several nuanced insights.

\medskip

\noindent\textbf{LLM comprehension of English as a matrix language is significantly disrupted by the introduction of elements from other languages.} Our experiments consistently show that inserting tokens from other languages—Arabic, Chinese, French, or German—into English text leads to a drop in LLM comprehension. This drop does not appear to stem solely from unfamiliarity with CSW, as similar performance declines were observed when randomly inserting foreign tokens (as in the ratio-token method from Experiment 2). Instead, these findings point to a more fundamental difficulty: LLMs struggle to process disrupted monolingual structures and integrate mixed linguistic signals effectively.
\noindent\textbf{Embedding English tokens into other languages often improves LLM comprehension of the original text.} LLMs frequently exhibited improved comprehension on non-English texts when English tokens were embedded, surpassing their baseline performance on the original monolingual versions of the same benchmarks.

\medskip

\noindent\textbf{Code-switching complexity does not linearly correlate with performance degradation.} In our "extreme" CSW experiments, increasing the number of embedded languages or mixing script types did not consistently lead to greater declines in model performance compared to simpler two-language settings. These findings suggest that degradation is not a direct function of multilingual complexity, but rather emerges from a nuanced interaction between specific language combinations and model-specific linguistic representations.

\medskip

\noindent\textbf{While prompting helps some models mitigate degradation, fine-tuning offers a more reliable solution.}
We evaluated two strategies for mitigating the effects of code-switching: prompt-based and model-based. Explicitly prepending instructions about upcoming code-switched input (Table \ref{tab:prompting_results_super_super_global}) proved effective for some architectures—most notably the \textit{Qwen} family. However, this strategy was less effective, or even detrimental, for others like \textit{Llama} and \textit{Mistral}, likely due to interference with their internal processing. For models that did not benefit from prompting, such as \textit{Llama 8B}, we explored direct instruction fine-tuning on code-switched data. This approach led to a more consistent improvement. As shown in Figure \ref{fig:mitigation-IFT}, \textit{Llama 8B}, which suffered performance drops under prompting, partially recovered its accuracy after instruction tuning—demonstrating that fine-tuning is a more promising path for improving LLM robustness to code-switching.
\medskip

These findings underscore that while LLMs exhibit impressive multilingual capabilities, CSW introduces specific comprehension challenges distinct from monolingual processing. The asymmetric impact of English as a matrix versus embedded language highlights areas requiring further research. While mitigation is possible, the model-specific nature of these solutions points towards the need for more adaptive approaches to ensure reliable LLM performance in real-world multilingual environments.

\section*{Limitations}
While our study adopts a controlled evaluation setup for both linguistically and non-linguistically motivated code-switching, the noun-token approach we employ reflects one of the fundamental forms of linguistically grounded, naturalistic switching. However, more complex forms of code-switching may induce more severe performance degradation. Future work should investigate how higher-complexity switching patterns affect LLMs' understanding.

Additionally, in our non-linguistically motivated ratio-token experiments, the substitution rate was fixed at 20\%. Exploring how variation in this ratio affects model behavior could yield a more nuanced understanding of the impact of non-linguistically grounded switching on LLM comprehension.

\bibliography{googlescholar,aclanthology}
\newpage
\newpage
\onecolumn
\appendix
\section{Additional Details}
All experiments were conducted using NVIDIA A100 (40GB VRAM) and A10 (24GB VRAM) GPU clusters. The compute allocation totaled 22 GPU-days, comprising 8 GPU-days on 8×A100 nodes and 14 GPU-days on 4×A10 nodes.
\section{Code-Switched Text Generation Approaches and Component Selection}
\label{app:generation_details}

This section details our selection process for model components used in generating code-switched (CSW) text, as introduced in Section~\ref{sec:methods}. Our objective was to identify the most effective LLM and alignment backbone for producing fluent, grammatically valid CSW outputs suitable for benchmark construction.

\subsection{LLM Selection for Generation}
\label{app:llm_selection_for_generation}

We compared Claude 3.5 Sonnet and GPT-4o as generation modules for both the Alignment-Based and LLM-Centric pipelines. For each matrix–embedded language pair (EN$\rightarrow$AR, ZH, FR, DE), we sampled 100 samples from the \textit{Belebele}, \textit{MMLU}, and \textit{XNLI} benchmarks. Both models generated noun-token CSW sentences under linguistically grounded prompting that adhered to the Equivalence Constraint Theory (ECT) and Matrix Language Frame (MLF) model.

Bilingual annotators conducted pairwise preference evaluations of the outputs, focusing on a single criterion: which code-switched sentence sounded more natural to them. Claude was consistently favored, with preference rates ranging from 52\% to 62\% across languages, as shown in Table~\ref{tab:llm_human_eval_appendix}. Accordingly, Claude was selected as the generation model for all subsequent CSW construction.

\begin{table}[H]
\centering
\begin{tabular}{l|c|c}
\toprule
Embedded Language & Claude (\%) & GPT-4o (\%) \\
\midrule
Arabic & 55 & 45 \\
Chinese & 57 & 43 \\
French & 52 & 48 \\
German & 62 & 38 \\
\bottomrule
\end{tabular}
\caption{Human preferences for CSW text generated by Claude vs. GPT-4o (100 examples per language pair).}
\label{tab:llm_human_eval_appendix}
\end{table}

\subsection{Embedding Backbone Selection}
\label{app:embedding_model_selection}

To identify the best embedding model for alignment in the Alignment-Based Pipeline, we evaluated AWESOME with mBERT (AWESOME's default embedding model) and LaBSE. For each language pair, 300 noun-token CSW sentences were generated using alignments from each configuration, with substitution handled by Claude.

Using GPT-4o as an LLM-based judge, we found that LaBSE-based alignments consistently yielded more natural and fluent code-switched outputs than those derived from mBERT, with clear preferences observed for Arabic (89.0\%), Chinese (91.3\%), and French (74.7\%). For German, the preference was more modest (55.3\%), though still in favor of LaBSE. GPT-4o was selected as the evaluator due to its strong multilingual capabilities and demonstrated aptitude in CSW understanding across typologically diverse languages. Importantly, using GPT-4o rather than Claude to evaluate outputs avoids the potential biases introduced by self-evaluation, such as output familiarity or training data memorization, thus providing a more neutral and reliable assessment of generation quality. Results presented in Table~\ref{tab:real_embedding_eval_appendix}, informed our decision to adopt LaBSE as the default embedding backbone for alignment in all subsequent experiments.

\begin{table}[H]
\centering
\begin{tabular}{l|c|c}
\toprule
Embedded Language & LaBSE (\%) & mBERT (\%) \\
\midrule
Arabic & 89.0 & 11.0 \\
Chinese & 91.3 & 8.7 \\
French & 74.7 & 25.3 \\
German & 55.3 & 44.7 \\
\bottomrule
\end{tabular}
\caption{GPT-4o preference rates for CSW text generated using LaBSE vs. mBERT alignments. Percentages reflect outcome ratios from 300 evaluation instances per language.}
\label{tab:real_embedding_eval_appendix}
\end{table}

\subsection{LLM-Centric Method Prompts}
\label{app:llm_centric_prompts}

The LLM-centric pipeline (Section \ref{sec:csw_generation_approaches}) uses a two-step prompting strategy:
\begin{enumerate}[nosep,leftmargin=1.5em]
  \item \textbf{Placeholder identification} — mark every switchable noun in the English sentence with a placeholder mask (\texttt{\#\#\#\#\#\#\#}).
  \item \textbf{Placeholder filling} — substitute each sentinel with the aligned word(s) from the parallel target-language sentence, yielding the final code-switched version.
\end{enumerate}

\begin{figure}[H]
  \centering
  \scriptsize
  \begin{minipage}{0.98\textwidth}
\begin{lstlisting}[basicstyle=\ttfamily\scriptsize,breaklines=true,frame=single]
You are an expert linguist and code-switching analyst. Based on the Equivalence
Constraint Theory and the Matrix Language Frame model, identify nouns in the
input English sentence that would serve as appropriate code-switching points.

- Input variable: text (a single English sentence)
- Task: Find every noun (as a free content morpheme) that can be switched under
  the theories above.
- Transformation: Replace each identified noun in the sentence with "#######".
- Output: Return only the transformed sentence with nouns replaced by "#######".
- The substituted words blend seamlessly into the text, following natural
  bilingual speech patterns.
- Adjust the target language words as needed (e.g., inflection, gender,
  number) so that the text remains syntactically correct.
- Ensure that nouns in common expressions are not code-switched.
- Don't return any summary or introduction, just the processed text

[English text]
{text}
\end{lstlisting}
  \caption{Step 1 — \emph{Placeholder identification} prompt (noun-token variant).}
  \label{fig:placeholder_id}
  \end{minipage}
\end{figure}
\begin{figure}[H]
  \centering
  \scriptsize
  \begin{minipage}{0.98\textwidth}
\begin{lstlisting}[basicstyle=\ttfamily\scriptsize,breaklines=true,frame=single]
You will be given a pair of parallel texts in English and {target_language}.

Your goal is to produce a code-switched version of the English text by replacing
each of the hashtag-sequences (#######) in the English text with their
{target_language} counterparts from the {target_language} text, ensuring that:
- The substituted words blend seamlessly into the text, following natural
  bilingual speech patterns.
- The text should be grounded with the principles of the Equivalence Constraint
  Theory and the Matrix Language Frame model.
- Adjust the target language words as needed (e.g., inflection, gender, number)
  so that the text remains syntactically correct.
- The original meaning and flow of the text are maintained.
- All the hashtag-sequences (#######) have to be replaced with text from the
  {target_language} text.
- Use only the words from the {target_language} text.
- Return only the code-switched text, without any additions or explanations.

[English text with placeholders]
{placeholder_text}

[{target_language} text]
{target_text}

[Code-switched English and {target_language}]
\end{lstlisting}
  \caption{Step 2 — \emph{Placeholder filling} prompt (noun-token variant).}
  \label{fig:placeholder_fill}
  \end{minipage}
\end{figure}

\begin{figure}[H]
  \centering
  \scriptsize
  \begin{minipage}{0.98\textwidth}
\begin{lstlisting}[basicstyle=\ttfamily\scriptsize,breaklines=true,frame=single]
You will be given an English sentence with placeholders (#######) and its
parallel sentence in {target_language}.
Replace each placeholder with the corresponding segment from the
{target_language} text, ensuring:
- The inserted text matches the target-language phrasing (inflections, gender,
  number).
- The final sentence reads naturally as mixed English and {target_language}.
- Preserve the original sentence order.
Return only the filled sentence, no extra comments.

[English with placeholders]
{placeholder_text}

[{target_language} parallel text]
{target_text}

[Mixed code-switched result]
\end{lstlisting}
  \caption{Prompt used in the \emph{ratio-token} variant (random placeholder insertion).}
  \label{fig:ratio_prompt}
  \end{minipage}
\end{figure}

\subsection{Final Generation Approach Selection}
\label{app:final_method_selection_concise}

We compared the Alignment-Based Pipeline and the LLM-Centric Method for generating noun-token CSW text across 100 samples per language and benchmark. Results are presented in Table \ref{tab:real_method_eval_appendix}. Pairwise evaluation via GPT-4o favored the LLM-Centric approach in all settings, with the strongest preferences for Chinese (66\%) and French (63.8\%). Based on these results, we adopt the LLM-Centric Method for all noun-token CSW benchmark construction, while retaining the Alignment-Based Pipeline for tasks requiring explicit control over substitution rates (e.g., ratio-token generation).

\begin{table}[H]
\centering
\begin{tabular}{l|c|c}
\toprule
Embedded Language & LLM-Centric (\%) & Alignment-Based (\%) \\
\midrule
Arabic & 56.1 & 43.9 \\
Chinese & 66.0 & 34.0 \\
French & 63.8 & 36.2 \\
German & 53.4 & 46.6 \\
\bottomrule
\end{tabular}
\caption{GPT-4o preferences between generation methods for noun-token CSW outputs.}
\label{tab:real_method_eval_appendix}
\end{table}
% \section{Prompt for LLM-as-a-Judge Evaluation of Code-Switching Generation Quality}
\label{app:llm_judge_prompt}
\begin{figure}[H]
\raggedleft
\begin{lstlisting}
You have two code-switched sentences, A and B, each blending English (matrix language) with {second_language}. Follow these steps and then choose the better sentence (A or B):

1. Assess Fluency: check which sentence flows most naturally, like plausible bilingual speech.
2. Assess Depth of Mixing: check which sentence meaningfully integrates both languages rather than inserting isolated tokens.
3. Assess Switch Grammar: check which sentence has grammatically valid switch points under Equivalence Constraint Theory.
4. Assess Consistency: check which sentence uses English as its grammatical frame and embeds {second_language} elements appropriately under the Matrix Language Frame model.
5. Assess Overall Coherence: check which sentence remains clear and plausible as a whole despite the language mixing.

After evaluating all five criteria, return A or B with no further explanation.

Sentences:
A: {sentence_one}
B: {sentence_two}

Output:
\end{lstlisting}
\caption{The prompt given to Claude 3.5 Sonnet for choosing the best summary between the baseline and LLM-generated summaries.}
\label{fig:method_selection_llm_judge_prompt}
\end{figure}

\section{Instructional Prompt for Prompt-Based Mitigation}
\label{app:prompt_mitigation}

\subsection*{\textit{Belebele} Prompt}
\begin{figure}[H]
\raggedleft
\begin{lstlisting}
You are an expert in understanding code-switched text. You will be given a passage and a question in code-switched English and Arabic. You have to understand them and respond to the given question with best answer: A, B, C, or D.
\end{lstlisting}
\caption{Instructional prompt prepended for \textit{Belebele} multiple-choice QA tasks.}
\label{fig:belebele_prompt}
\end{figure}

\subsection*{\textit{MMLU} Prompt}
\begin{figure}[H]
\raggedleft
\begin{lstlisting}
You are an expert in understanding code-switched text. You will be given a question in code-switched English and Arabic. You have to understand it and respond to the given question with best answer: A, B, C, or D.
\end{lstlisting}
\caption{Instructional prompt prepended for \textit{MMLU} multiple-choice QA tasks.}
\label{fig:mmlu_prompt}
\end{figure}

\subsection*{\textit{XNLI} Prompt}
\begin{figure}[H]
\raggedleft
\begin{lstlisting}
You are an expert in understanding code-switched text. You will be given two code-switched passages that correspond to a premise and a hypothesis in code-switched English and Arabic text. You have to understand them and respond with the best answer: 0, 1, or 2.
\end{lstlisting}
\caption{Instructional prompt prepended for \textit{XNLI} natural language inference tasks.}
\label{fig:xnli_prompt}
\end{figure}

\section{Instruction Tuning for Model-Based Mitigation}
\label{app:ift_details_full}

We fine-tuned \textit{LLaMA-3.1-8B-Instruct} to improve its comprehension of code-switched text using a targeted instruction-tuning dataset. Full-model training was conducted over a single epoch using a learning rate of $2 \times 10^{-6}$ with linear decay and 5\% warmup. Training leveraged mixed-precision BF16 and dynamic sequence packing within a 4096-token window, and a batch-size of four.

\subsection{Dataset Preparation}
\label{app:dataset_prep_templating}

The training data was derived from parallel TED Talk translations \citep{qi-etal-2018-pre}, selecting English sentences longer than 70 words and their Arabic, Chinese, French, and German equivalents. Each English sentence was converted into four code-switched variants using the LLM-Centric Method (Appendix~\ref{app:final_method_selection_concise}). The final dataset included over 14,000 examples, shuffled and formatted as instruction–response pairs.

\subsection{Prompt Templates for Instruction Tuning}
\label{app:prompt_templates_sft}

To prevent overfitting to fixed phrasing, each training instance was paired with a randomly selected prompt from a pool of five semantically equivalent instruction templates. These templates varied in their surface structure but uniformly instructed the model to blend the matrix English sentence with embedded nouns from the translation. Figures~\ref{fig:ipt1}–\ref{fig:ipt5} illustrate the five styles used.

\begin{figure}[H]
\centering
\begin{lstlisting}[language=SQL, basicstyle=\ttfamily\footnotesize, breaklines=true, frame=single]
Take this English sentence and infuse it with <LANGUAGE> code-switching:
English: "<ENGLISH_SENTENCE>"
<LANGUAGE>: "<TRANSLATION_SENTENCE>"
\end{lstlisting}
\caption{Infusion-style template.}
\label{fig:ipt1}
\end{figure}

\begin{figure}[H]
\centering
\begin{lstlisting}[language=SQL, basicstyle=\ttfamily\footnotesize, breaklines=true, frame=single]
Convert the following English line into a code-switched mix with <LANGUAGE>:
English: "<ENGLISH_SENTENCE>"
<LANGUAGE>: "<TRANSLATION_SENTENCE>"
\end{lstlisting}
\caption{Conversion-style template.}
\label{fig:ipt2}
\end{figure}

\begin{figure}[H]
\centering
\begin{lstlisting}[language=SQL, basicstyle=\ttfamily\footnotesize, breaklines=true, frame=single]
Blend English and <LANGUAGE> in the sentence below:
English text: "<ENGLISH_SENTENCE>"
<LANGUAGE> equivalent: "<TRANSLATION_SENTENCE>"
\end{lstlisting}
\caption{Blending-style template.}
\label{fig:ipt3}
\end{figure}

\begin{figure}[H]
\centering
\begin{lstlisting}[language=SQL, basicstyle=\ttfamily\footnotesize, breaklines=true, frame=single]
Generate a code-switched rendition by swapping in <LANGUAGE>:
English original: "<ENGLISH_SENTENCE>"
<LANGUAGE> snippet: "<TRANSLATION_SENTENCE>"
\end{lstlisting}
\caption{Rendition-style template.}
\label{fig:ipt4}
\end{figure}

\begin{figure}[H]
\centering
\begin{lstlisting}[language=SQL, basicstyle=\ttfamily\footnotesize, breaklines=true, frame=single]
Switch parts of this English sentence into <LANGUAGE>:
English: "<ENGLISH_SENTENCE>"
<LANGUAGE>: "<TRANSLATION_SENTENCE>"
\end{lstlisting}
\caption{Switching-style template.}
\label{fig:ipt5}
\end{figure}
\end{document}